\documentclass[conference]{IEEEtran}
\IEEEoverridecommandlockouts
\usepackage{cite}
\usepackage{amsmath,amssymb,amsfonts}
\usepackage{algorithmic}
\usepackage{graphicx}
\usepackage{textcomp}
\usepackage{xcolor}
\usepackage{hyperref}
\usepackage{url}
\usepackage{microtype}      
\usepackage{xcolor}         
\usepackage{amsmath}
\usepackage{algorithm}
\usepackage{algorithmic}
\usepackage{mathabx}
\usepackage{amssymb}
\usepackage{multirow}
\usepackage{adjustbox}
\usepackage{tabularx}
\usepackage{caption}
\usepackage{graphicx}
\usepackage{subcaption}
\usepackage{setspace}
\usepackage{graphicx}
\usepackage{etoolbox}
\def\BibTeX{{\rm B\kern-.05em{\sc i\kern-.025em b}\kern-.08em
    T\kern-.1667em\lower.7ex\hbox{E}\kern-.125emX}}



\begin{document}

\title{RetCompletion:High-Speed Inference Image Completion with Retentive Network}

\author{
    \IEEEauthorblockN{
        Yueyang Cang\textsuperscript{\dag},
        Pingge Hu\textsuperscript{\ddag},
        Xiaoteng Zhang\textsuperscript{\dag},
        Xingtong Wang\textsuperscript{\dag},
        Yuhang Liu\textsuperscript{\dag},
        Li Shi\textsuperscript{\dag,*}
    }
    \IEEEauthorblockA{
        \textsuperscript{\dag} Tsinghua University
    }
    \IEEEauthorblockA{
        \textsuperscript{\ddag} China Academy of Information and Communications Technology
    }
    \thanks{* Corresponding author}
}

\maketitle

\begin{abstract}
Time cost is a major challenge in achieving high-quality pluralistic image completion. Recently, the Retentive Network (RetNet) in natural language processing offers a novel approach to this problem with its low-cost inference capabilities. Inspired by this, we apply RetNet to the pluralistic image completion task in computer vision. We present RetCompletion, a two-stage framework. In the first stage, we introduce Bi-RetNet, a bidirectional sequence information fusion model that integrates contextual information from images. During inference, we employ a unidirectional pixel-wise update strategy to restore consistent image structures, achieving both high reconstruction quality and fast inference speed. In the second stage, we use a CNN for low-resolution upsampling to enhance texture details. Experiments on ImageNet and CelebA-HQ demonstrate that our inference speed is 10$\times$ faster than ICT and 15$\times$ faster than RePaint. The proposed RetCompletion significantly improves inference speed and delivers strong performance. 
\end{abstract}


\section{Introduction}
\label{sec:intro}

Pluralistic image completion, also known as image inpainting, is a crucial research area with various applications, including object removal and photo restoration \cite{barnes2009patchmatch,criminisi2004region,dale2009image,wan2020bringing}. CNN-based methods \cite{CNN3} have demonstrated impressive results by capturing local texture patterns, but they often struggle to model global structures, leading to suboptimal image reconstruction quality. To overcome this limitation, researchers have introduced hybrid models combining Transformers and CNNs \cite{wan2021high,zheng2022bridging,li2022mat}. While these approaches significantly improve reconstruction quality and produce diverse results by modeling the underlying data distribution, Transformer-based pixel-wise generation involves extensive feature fusion calculations. This computational overhead increases inference time, limiting the practicality of these methods, especially in real-time applications. Therefore, developing algorithms that maintain high-quality reconstruction while improving computational efficiency remains a critical challenge in this domain.

Recently, Retentive Network (RetNet) \cite{sun2023retentive} has shown substantial potential in natural language processing due to its multi-scale retention mechanism, which bridges parallel training and recurrent inference. This capability enables RetNet to process information efficiently, even in pixel-wise generation tasks. However, applying RetNet directly to vision tasks presents challenges, as image information is not unidirectional like language data.

In this work, we propose RetCompletion, a novel image completion framework designed to address the challenges of slow inference and inconsistent image reconstruction. RetCompletion operates in two stages: the first stage leverages a Bi-RetNet architecture for low-resolution pixel-wise image generation, while the second stage uses a CNN for high-resolution texture refinement. Extensive experiments on datasets such as ImageNet and CelebA-HQ demonstrate that RetCompletion significantly accelerates inference while maintaining high reconstruction quality.

The key contributions of this work are:
\begin{enumerate}
    \item \textbf{First application of RetNet to image completion}: We introduce RetNet for the first time in image completion tasks, utilizing its parallel training and recursive inference to accelerate the process.
    \item \textbf{Bi-RetNet with bidirectional fusion}: Our Bi-RetNet architecture fuses forward and backward contextual information, improving consistency and realism, particularly when reconstructing large masked areas.
    \item \textbf{Efficient pixel-wise inference based on RetNet}: RetCompletion's pixel-wise inference strategy, enabled by RetNet, is significantly faster than Transformer-based methods and produces better overall results by incorporating previously generated pixel information during inference.
\end{enumerate}

\section{Related Work}

\paragraph{Pluralistic Image Completion} The significance of Pluralistic Image Completion lies in providing a diverse approach to image processing, allowing for the creation of images with different styles and effects, enriching the toolbox in creative and design fields, and supporting diverse choices in decision-making processes.  PIC~\cite{CNN4} employs a dual-path framework based on probabilistic principles: one is the reconstructive path, which utilizes the given ground truth to obtain prior information about the missing parts and reconstructs the original image from this distribution. The other is the generative path, where the conditional prior is coupled with the distribution from the reconstructive path. ICT~\cite{wan2021high}  directly optimizes the log-likelihood in the discrete space in the first transformer-based stage without the need for additional assumptions. RePaint~\cite{lugmayr2022repaint} applies the Diffusion model to the image inpainting task, using a pre-trained unconditional DDPM~\cite{ho2020denoising} as the generative prior and modifying the reverse diffusion iterations by sampling the unmasked regions from the given image information. Since this technique doesn't alter or condition the original DDPM~\cite{ho2020denoising} network itself, the model can generate high-quality and diverse output images for any inpainting scenario.

\paragraph{Retentive Network} Retentive Network~\cite{sun2023retentive} introduces the retention mechanism with a dual form of recurrence and parallelism. It has three computation paradigms,i.e., parallel, recurrent, and chunkwise recurrent.  We can train parallelly by using parallel paradigm while conducting inference recurrently using recurrent and chunkwise paradigms. The retention mechanism utilizes a rotation-based positional encoding along with a decay term to effectively model the position information of tokens, known as xPos~\cite{sun2022length}, a relative position embedding proposed for Transformer. We attempt to extend this method to two-dimensional images.
\section{Methods}

\begin{figure*}
    \centering
    \includegraphics[width=1\linewidth]{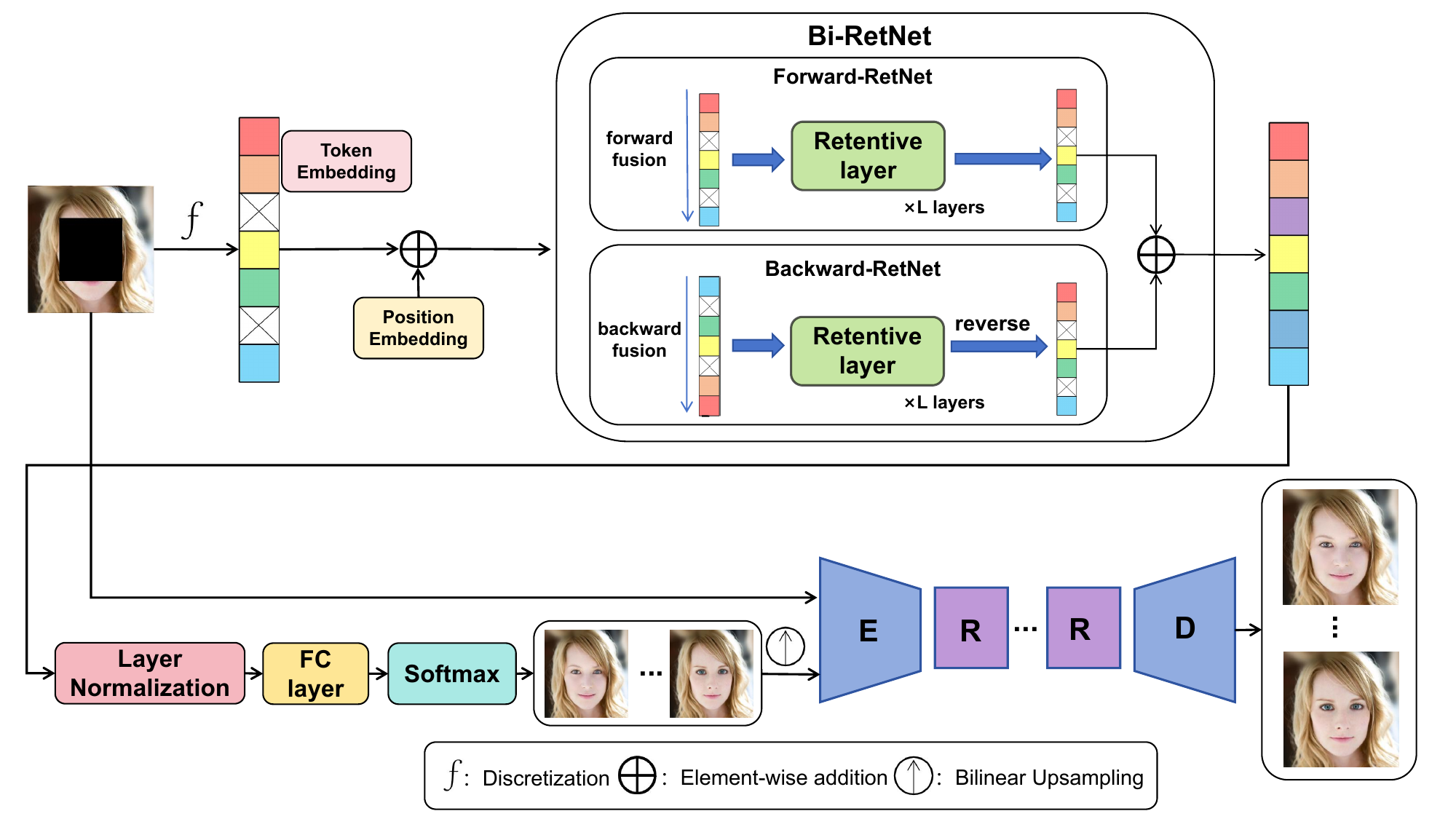}
    \caption{\textbf{Pipeline Overview.} Our method consists of two networks, which are trained separately. Based on the Bi-RetNet, the first network is employed for completing low-dimensional images. A parallel representation is utilized during training, predicting all pixels simultaneously to expedite the training process. In contrast, during inference, a recurrent representation is employed, predicting one pixel at a time to enhance the quality of the generated image. The second network, built on a CNN architecture, comprises an encoder, a decoder, and multiple residual blocks. Its primary function is to restore high-dimensional images from their low-dimensional counterparts.}
    \label{pipeline}
\end{figure*}

The overall pipeline of our method can be seen in Figure. \ref{pipeline}, which consists of two stages. The first stage is utilized to complete low-resolution images based on Bi-RetNet, while the second stage generates high-resolution images based on CNN.

\subsection{Preprocessing}

To reduce the computational cost of attention calculations during preprocessing, we first downsample the input image from its original resolution $H \times W$ to a lower-resolution version $L \times L$:

\begin{equation} \label{eq3} \bar{I}{L \times L \times 3} = \text{Downsampling}(I{H \times W \times 3}) \end{equation}

This step simplifies the image, reducing the number of pixels that need to be processed during subsequent stages.

RGB color channels typically exist in a high-dimensional space ($256^3$ colors), which makes direct processing computationally expensive. To handle this, we generate a compact visual vocabulary by applying K-Means clustering on the entire ImageNet dataset \cite{russakovsky2015imagenet}, reducing the space to 512 representative colors. For each unmasked pixel, we map its color to the nearest representative color from this vocabulary. The image is then raster-scanned and reshaped into a sequence, which is necessary for the RetNet model:

\begin{equation}\label{eq4} S_{L^2 \times 1} = \text{reshape}(\text{project}(\bar{I}_{L \times L \times 3})) \end{equation}

We also create a binary mask sequence, where 1 indicates a masked pixel and 0 represents an unmasked pixel:

\begin{equation}\label{eq5}
M_{L^2 \times 1} = \left\{\mathbb{I}(S_i \text{ is masked}), i=1,2,\ldots,L^2\right\}
\end{equation}

\paragraph{Feature Encoding} To convert each pixel's color into a feature vector, we use a trainable embedding. This transforms the discrete color values from the visual vocabulary into $d$-dimensional feature vectors, which will serve as inputs to RetNet.

\paragraph{Position Encoding} RetNet uses positional encoding to track the location of tokens in a sequence. For 2D images, we introduce a learnable position embedding that captures spatial information. This embedding, combined with the feature encoding, forms the input sequence for RetNet:

\begin{equation}\label{eq6} X_{L^2 \times d} = \text{FE}(S_{L^2 \times 1} \odot M_{L^2 \times 1}) + \text{PE}_{L^2 \times d} \end{equation}

By combining the color features and positional information, this sequence serves as the input to the RetNet model, allowing it to process the image efficiently in the subsequent stages.

\subsection{Appearance Priors Reconstruction by Bi-RetNet}

Traditional RetNet operates with unidirectional information flow, which is suitable for natural language processing. However, image completion requires integrating contextual information from multiple directions. To address this, we developed a bidirectional RetNet model consisting of a Multi-Head Forward-RetNet and a Multi-Head Backward-RetNet. These two RetNets share the same structure but have different parameters, enabling them to capture information from different directions.

\paragraph{Multi-Head Forward-RetNet} In this model, we utilize $h$ heads, where each head has a feature dimension of $d_{\text{head}} = d/h$. Different heads use different parameter matrices $W_Q, W_K, W_V \in \mathbb{R}^{d_{\text{head}} \times d_{\text{head}}}$. The retention for each head is computed as:

\begin{equation} head_i = Retention(X,W_i) \end{equation}

The multi-head outputs are concatenated and normalized using GroupNorm:

\begin{equation} Y = GroupNorm_h(Concat(head_1, \ldots, head_h)) \end{equation}

The final output for each forward layer is computed as:

\begin{equation} Y^l_{forward} = MSR(LN(X^l_{forward})) + X^l_{forward} \end{equation} \begin{equation} X^{l+1}{forward} = FFN(LN(Y^l{forward}))+Y^l_{forward} \end{equation} where $X^1$ is the sequence obtained from the preprocessing stage, and $l$ denotes the layer index.

\paragraph{Multi-Head Backward-RetNet} The Multi-Head Backward-RetNet follows the same procedure, except that it processes the reversed input sequence. After computation, the result is reversed back to its original order, yielding $X^{(L+1)}_{backward}$.

\paragraph{Feature Fusion}

To combine the forward and backward information, we perform feature fusion by adding the outputs from the forward and backward passes. Layer normalization, fully connected layers, and softmax are then applied to produce a per-pixel distribution of 512 possible colors:

\begin{equation} P(x|X, \theta) = softmax(FC(LN(X^{L+1}{forward}+X^{L+1}{backward}))) \end{equation}

This fusion of forward and backward information allows the model to capture richer contextual dependencies, leading to more accurate and coherent image reconstructions.

\paragraph{Loss Function}

Similar to BERT \cite{devlin2018bert}, we employ the Masked Language Model (MLM) objective to optimize the RetNet. The loss function minimizes the negative log-likelihood of the masked pixels, ensuring that the model learns to predict missing regions accurately:

\begin{equation} L_\text{MLM} = \underset{X}{\mathbb{E}}[\frac{1}{N}\sum_{n=1}^{N}-\log p(x_{n}|X,\theta)] \end{equation} where $N$ represents the number of masked pixels in the image. By minimizing this loss, the generated images approach the ground truth, resulting in high-quality reconstructions.

\subsection{Parallel Training}

During training, we utilize both the parallel and chunkwise recurrent representations to accelerate the process. Specifically, we choose to predict all masked pixels simultaneously, rather than sequentially, in order to improve training efficiency.

In this approach, masked pixels receive color information exclusively from unmasked pixels, meaning that masked pixels do not incorporate information predicted for other masked pixels, even those earlier in the sequence. This strategy allows for more efficient computation, as it reduces dependencies between predictions and enables faster iteration over large datasets.

By employing this parallelized method, we are able to reduce the overall training time, especially when handling high-dimensional data.

\subsection{Pixel-wise Inference}

In the inference stage, we adopt a pixel-wise inference method, which has proven to be significantly more effective compared to predicting all pixels simultaneously. The pixel-wise approach allows the model to incorporate newly predicted pixel information step by step, improving the overall quality of the generated images. This advantage is made possible by the Bi-RetNet architecture, which enables fast updates during inference, a capability that Transformer-based models lack.

We begin by performing information fusion on the initial image to generate integrated representations, $S_{forward}$ and $S_{backward}$. Then, we predict the masked pixels one by one in a raster-scan manner. At each step, we update the retention state of the forward RetNet with the new pixel information, ensuring that each subsequent pixel prediction benefits from previous predictions.

The inference process is detailed in Algorithm~\ref{alg1}, where the model integrates the forward and backward information for each pixel prediction and updates the RetNet's retention state after each step.

\begin{algorithm}
    \renewcommand{\algorithmicrequire}{\textbf{Input:}}
    \renewcommand{\algorithmicensure}{\textbf{Output:}}
    \caption{Pixel-wise Inference}
    \label{alg1}
    \setstretch{1.1}
    \scriptsize  
    \begin{algorithmic}[1]
        \STATE \textbf{Initialization:}
        \STATE Compute initial $Q_0 = X_0 W_Q$, $K_0 = X_0 W_K$, $V_0 = X_0 W_V$
        \STATE Initialize $S_{forward}$ and $S_{backward}$ with $Q_0, K_0, V_0$
        
        \STATE \textbf{Pixel-wise Inference:}
        \FOR{$n = 1$ to $N$ (where $j_n$ are the indices of masked pixels)}
            \STATE Retrieve positional encodings: $\bar{X}_{forward,n1} = PE_{j_n}, \bar{X}_{backward,n1} = PE_{j_n}$
            \FOR{each layer $l$ from $1$ to $L$}
                \STATE \# Forward Pass
                \STATE $Y_{forward,nl} = MSR(LN(\bar{X}_{forward,nl})) + \bar{X}_{forward,nl}$
                \STATE $\bar{X}_{forward,n(l+1)} = FFN(LN(Y_{forward,nl})) + Y_{forward,nl}$
                \STATE \# Backward Pass
                \STATE $Y_{backward,nl} = MSR(LN(\bar{X}_{backward,nl})) + \bar{X}_{backward,nl}$
                \STATE $\bar{X}_{backward,n(l+1)} = FFN(LN(Y_{backward,nl})) + Y_{backward,nl}$
            \ENDFOR
            \STATE \# Combine forward and backward results for prediction
            \STATE $P(x_n) = \text{softmax}(FC(LN(\bar{X}_{forward,n(L+1)}+\bar{X}_{backward,n(L+1)})))$
            \STATE Sample pixel value: $x_n \sim P(x_n)$
            \STATE Update pixel embedding: $X_n = FE(x_n) + PE_{j_n}$
            \STATE \# Update forward RetNet state
            \STATE Compute new $Q_n = X_n W_Q$, $K_n = X_n W_K$, $V_n = X_n W_V$
            \STATE Update $S_{forward}$ with $Q_n, K_n, V_n$
        \ENDFOR
    \end{algorithmic}
\end{algorithm}

\subsection{Guided Upsampling}

ollowing ICT~\cite{wan2021high}, we use a CNN-based guided upsampling network to reconstruct the high-resolution image. After reshaping the appearance priors into a low-resolution image, we first apply bilinear interpolation to upscale it to the original resolution. The upsampled image, along with the original image and mask, is then fed into the upsampling network. We train the network using both $L_1$ loss and adversarial loss, with the upsampling network and discriminator being optimized jointly.

\section{Experiments}

\begin{figure*}
    \centering
    \includegraphics[width=1\linewidth]{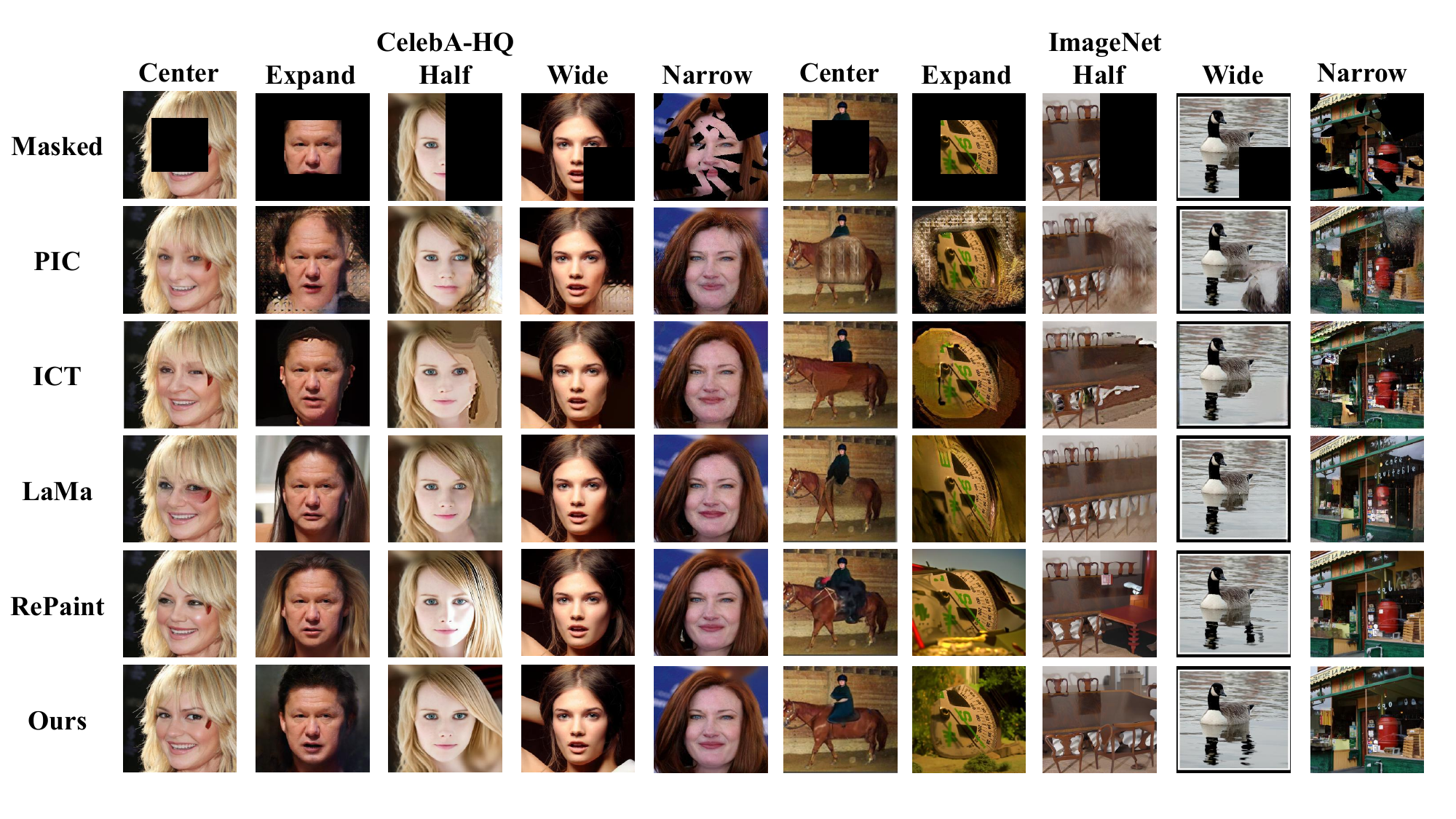}
    \caption{\textbf{Sample images for user study.}}
    \label{user_img}
\end{figure*}

In our experiments, we evaluate the performance of the proposed method using two datasets: CelebA-HQ~\cite{karras2017progressive} and ImageNet~\cite{russakovsky2015imagenet}. We perform both quantitative and qualitative evaluations to assess the quality of image completion and the inference speed. Quantitative comparisons are conducted against other state-of-the-art methods in terms of image quality and computational efficiency, while qualitative comparisons are based on user feedback. Note that all qualitative and quantitative results reported in this paper are based on a fixed image resolution of 256 pixels.

\subsection{Implementation Details}

To ensure fair comparisons across different datasets and methods, we follow the same configuration as ICT~\cite{wan2021high} for all experiments, as shown in Table~\ref{table1}. For the CelebA-HQ~\cite{karras2017progressive} and ImageNet~\cite{russakovsky2015imagenet} datasets, we retain the original training and test splits. Additionally, we employ PConv \cite{liu2018image} to generate diverse masks during training to simulate various image occlusion scenarios.

\begin{table}
    \centering  \captionsetup{justification=raggedright,singlelinecheck=false}
    \renewcommand\arraystretch{1.2}
    \begin{tabular}{c|c|c|c|c}
        \hline
        Datasets & h & d & N & $\mathbb{L}$ \\
        \hline
        CelebA-HQ~\cite{karras2017progressive} & 8 & 512 & 30 & 48$\times$48 \\
        ImageNet~\cite{russakovsky2015imagenet} & 8 & 1024 & 35 & 32$\times$32 \\
        \hline
    \end{tabular}
    \caption{\textbf{Retention Network parameter setting across different experiment}. h: Head number. d: The dimension of embedding space. N: Number of retention layers. \textbf{L}: The length of appearance prior.}
    \label{table1} 
\end{table}

\begin{table}
    \centering
    \resizebox{\linewidth}{!}{
        \renewcommand\arraystretch{1.2}
        \begin{tabular}{c|c|c|c|c|c|c|c}
            \hline
            \multicolumn{2}{c}{Dataset} & \multicolumn{3}{c}{CelebA-HQ~\cite{karras2019style}} & \multicolumn{3}{c}{ImageNet\cite{russakovsky2015imagenet}} \\
            \hline
            Method & Mask Ratio & PSNR & SSIM & LPIPS & PSNR & SSIM & LPIPS \\
            \hline
            PIC~\cite{CNN4}&\multirow{7}{*}{Wide} & 23.781 & 0.883 & 0.164 & 23.765 & 0.819 & 0.234 \\
            LaMa~\cite{suvorov2022resolution} && 27.581 & 0.928 & 0.045 & 26.099 & 0.865 & 0.105 \\
            RePaint~\cite{lugmayr2022repaint} && 27.496 & 0.931 & 0.059 & 25.768 & 0.859 & 0.134 \\
            $ICT^*$~\cite{wan2021high}  && 26.897 & 0.922 & 0.069 & 25.545 & 0.848 & 0.125 \\
            ICT~\cite{wan2021high}  && 27.139 & 0.932 & 0.063 & 25.886 & 0.862 & 0.107  \\
            $Ours^*$ && 27.643 & 0.926 & 0.053 & 25.989 & 0.852 & 0.118  \\
            Ours && 27.966 & 0.938 & 0.042 & 26.087 & 0.869 & 0.103 \\
            \hline
            PIC~\cite{CNN4} &\multirow{7}{*}{Narrow}& 25.823 & 0.901 & 0.062 & 24.091 & 0.823 & 0.098  \\
            LaMa~\cite{suvorov2022resolution} && 28.684 & 0.942 & 0.028 & 26.892 & 0.902 & 0.061  \\
            RePaint~\cite{lugmayr2022repaint} && 28.547 & 0.938 & 0.028 & 26.908 & 0.906 & 0.064 \\
            $ICT^*$~\cite{wan2021high}  && 28.242 & 0.932 & 0.041 & 26.887 & 0.898 & 0.079  \\
            ICT~\cite{wan2021high}  && 28.551 & 0.944 & 0.036 & 26.902 & 0.903 & 0.073  \\
            $Ours*$ && 28.397 & 0.935 & 0.031 & 26.882 & 0.901 & 0.071  \\
            Ours && 28.692 & 0.943 & 0.029 & 26.911 & 0.902 & 0.065  \\
            \hline
            PIC~\cite{CNN4} &\multirow{7}{*}{Half}& 21.484 & 0.852 & 0.238 & 19.498 & 0.708 & 0.354  \\
            LaMa~\cite{suvorov2022resolution} && 25.208 & 0.905 & 0.138 & 23.513 & 0.756 & 0.254 \\
            RePaint~\cite{lugmayr2022repaint} && 24.846 & 0.902 & 0.165 & 23.498 & 0.762 & 0.304  \\
            $ICT^*$~\cite{wan2021high}  && 24.356 & 0.896 & 0.179 & 23.476 & 0.748 & 0.278  \\
            ICT~\cite{wan2021high}  && 24.798 & 0.906 & 0.166 & 23.502 & 0.753 & 0.255  \\
            $Ours^*$ &&24.798 & 0.898 & 0.153 & 23.496 & 0.746 & 0.278  \\
            Ours &&25.103 & 0.907 & 0.145 & 23.512 & 0.759 & 0.262 \\
            \hline
            PIC~\cite{CNN4} &\multirow{7}{*}{Center}& 25.580 & 0.887 & 0.153 & 23.806 & 0.816 & 0.167  \\
            LaMa~\cite{suvorov2022resolution} && 28.529 & 0.940 & 0.039 & 26.276 & 0.886 & 0.086 \\
            RePaint~\cite{lugmayr2022repaint} && 28.556 & 0.940 & 0.041 & 26.304 & 0.886 & 0.093  \\
            $ICT^*$~\cite{wan2021high}  && 28.409 & 0.935 & 0.058 & 26.198 & 0.879 & 0.103  \\
            ICT~\cite{wan2021high}  && 28.496 & 0.942 & 0.052 & 26.282 & 0.888 & 0.096  \\
            $Ours^*$ &&28.504 & 0.932 & 0.045 & 26.245 & 0.880 & 0.092  \\
            Ours &&28.559 & 0.938 & 0.037 & 26.311 & 0.890 & 0.083 \\
            \hline
            PIC~\cite{CNN4} &\multirow{7}{*}{Expand}& 18.893 & 0.798 & 0.576 & 17.364 & 0.652 & 0.712  \\
            LaMa~\cite{suvorov2022resolution} && 23.382 & 0.878 & 0.342 & 20.384 & 0.697 & 0.534 \\
            RePaint~\cite{lugmayr2022repaint} && 23.376 & 0.882 & 0.435 & 20.439 & 0.702 & 0.629  \\
            $ICT^*$~\cite{wan2021high}  && 23.298 & 0.876 & 0.446 & 20.126 & 0.683 & 0.562  \\
            ICT~\cite{wan2021high}  && 23.379 & 0.879 & 0.432 & 20.324 & 0.698 & 0.544  \\
            $Ours^*$ && 23.339 & 0.872 & 0.398 & 20.218 & 0.696 & 0.541  \\
            Ours && 23.380 & 0.881 & 0.372 & 20.423 & 0.706 & 0.536 \\
            \hline
        \end{tabular}
    }
    \captionsetup{justification=raggedright,singlelinecheck=false}
    \caption{\textbf{Quantitative results on CelebA-HQ~\cite{karras2017progressive} and ImageNet~\cite{russakovsky2015imagenet} datasets with different mask types.} All the pluralistic image completion methods use Top-1 sampling. The models with $*$ indicate the prediction method that uses all pixel points simultaneously, while the models without $*$ indicate the prediction method that uses pixel-by-pixel prediction.
 }
    \label{table2}
\end{table}

\subsection{Quantitative Comparisons}
We quantitatively compare our method against several state-of-the-art (SOTA) image completion techniques using peak signal-to-noise ratio (PSNR), structural similarity index (SSIM), and learned perceptual image patch similarity (LPIPS). Experiments are conducted on CelebA-HQ~\cite{karras2017progressive} and ImageNet~\cite{russakovsky2015imagenet} datasets with five distinct mask types to assess performance across different occlusion patterns. For all pluralistic image completion methods, Top-1 sampling is applied during testing.

The results of the quantitative experiments are presented in Table \ref{table2}, which shows performance for the Wide and Expand mask types. Our method consistently outperforms most existing SOTA methods across various mask types and datasets, demonstrating clear superiority in terms of image quality and perceptual similarity, as evidenced by improvements in PSNR, SSIM, and LPIPS metrics. A notable observation is the significant difference between pixel-wise inference and simultaneous pixel estimation (denoted with * in the table). Our pixel-wise inference approach performs better, especially in scenarios with complex occlusions. We employ five distinct mask types in our experiments: Wide and Narrow masks (adapted from LaMa~\cite{suvorov2022resolution}), Half mask (random occlusion of parts of the image), Center mask (covering a central $64 \times 64$ region), and Expand mask (occluding everything except the central region). For the complete results across all mask types, please refer to Appendix.

\subsection{User study} 
To enhance the assessment of subjective quality, we additionally perform a user study to compare our method against other baseline approaches. We randomly select 50 images and apply various masks to each. Employing different image completion methods, including pluralistic image completion methods, we consistently used the Top-1 sampling result. Specifically, we present a set of five images generated by MED~\cite{liu2020rethinking}, PIC~\cite{CNN4}, EC~\cite{CNN5}, ICT~\cite{wan2021high}, and our method for each image. Users are then asked to rank the top three images that appear most realistic. Finally, we calculate the percentage of times each method ranked within the top three. Sample images for user study are shown in Figure. ~\ref{user_img}.

\begin{figure*}[htbp]
    \centering
    \begin{subfigure}[b]{0.48\linewidth}
        \centering
        \includegraphics[width=\linewidth]{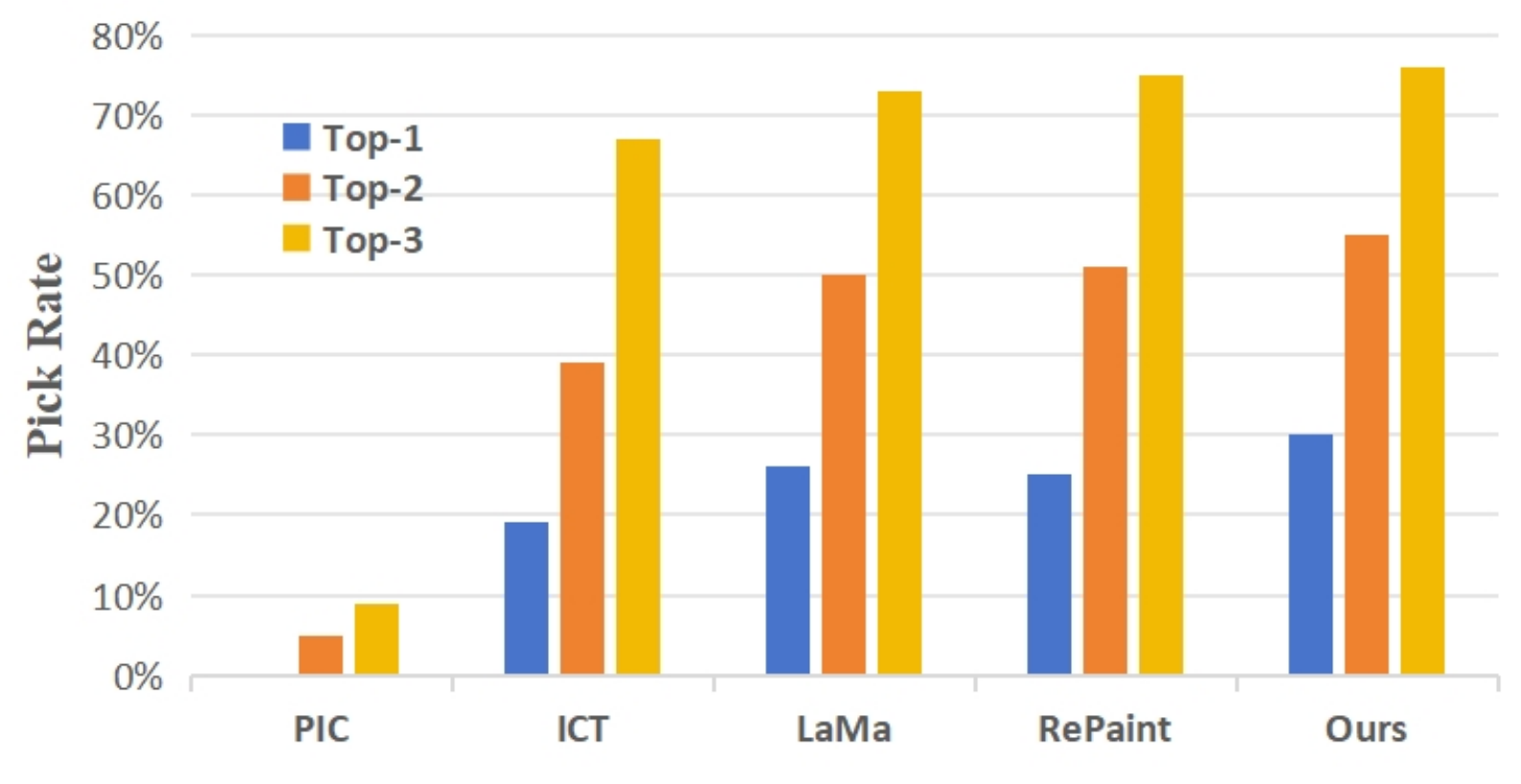}
        \caption{\textbf{Results of user study}}
        \label{user}
    \end{subfigure}
    \hfill
    \begin{subfigure}[b]{0.48\linewidth}
        \centering
        \includegraphics[width=\linewidth]{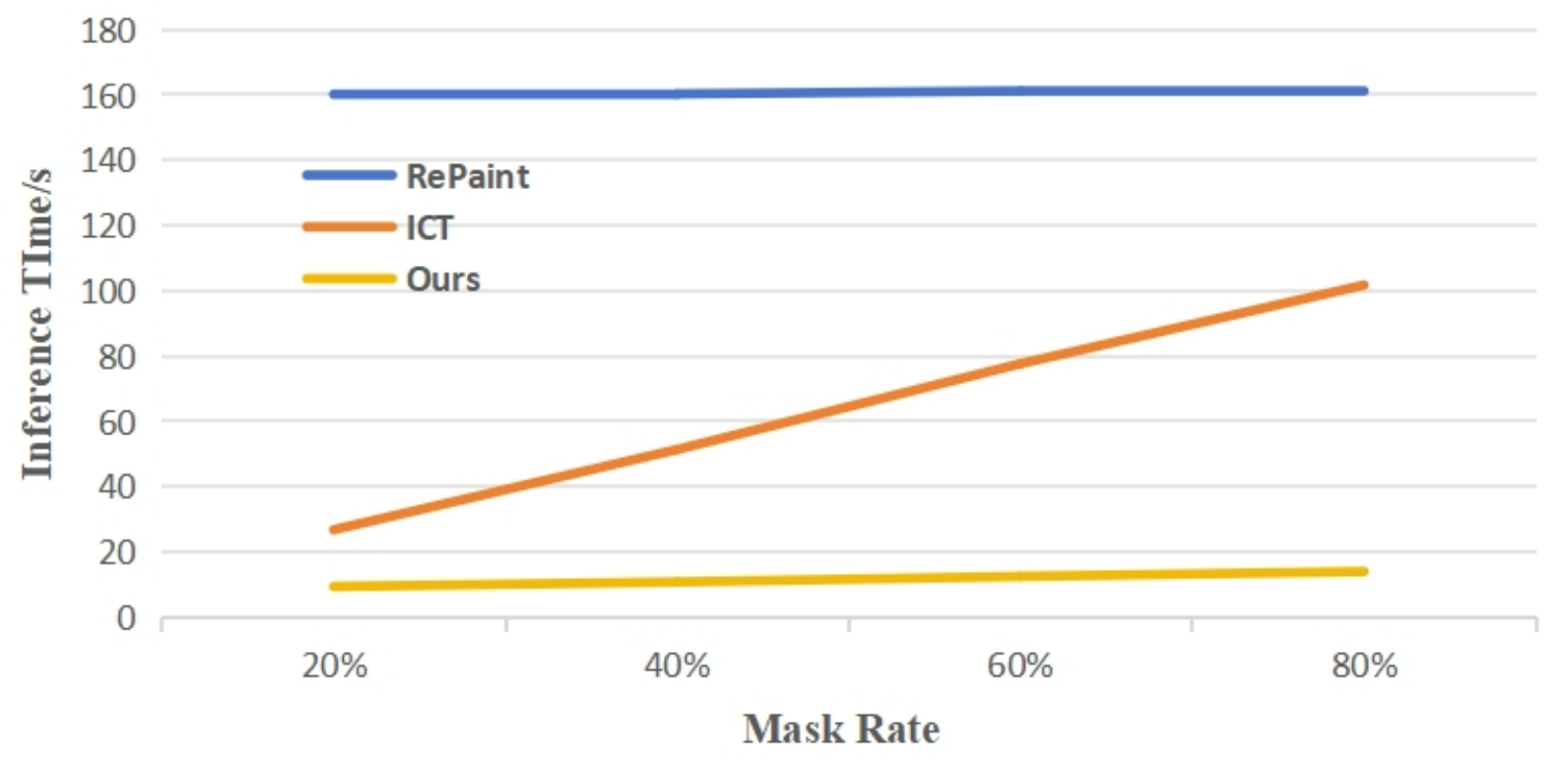}
        \caption{\textbf{Inference time comparison}}
        \label{time}
    \end{subfigure}
    \caption{Comparison of user study results and inference time.}
    \label{fig:comparison}
\end{figure*}

The results obtained from 200 users are shown in Figure~\ref{user}. The results show that our method significantly outperforms the PIC method in terms of visual quality. Additionally, our method shows a slight advantage over the ICT\cite{wan2021high}, LaMa~\cite{suvorov2022resolution}, and RePaint\cite{lugmayr2022repaint} methods. This demonstrates the superiority of our method in terms of visual perception.

\subsection{Inference Time}
To ensure a fair comparison, we measured the pixel-wise completion inference time of ICT~\cite{wan2021high}, RePaint~\cite{lugmayr2022repaint}, and our method on the ImageNet~\cite{russakovsky2015imagenet} dataset using a GeForce RTX 4090 GPU. The results, shown in Figure~\ref{time}, indicate that our method achieves significantly lower inference times, especially at higher mask rates.

RePaint maintains consistent inference time but requires numerous denoising iterations, leading to a higher overall time. ICT~\cite{wan2021high} recalculates attention for each pixel, causing a linear increase in time as the mask rate rises. In contrast, our method uses a recurrent computation paradigm, where only changed pixels are updated, resulting in minimal increases in inference time and consistently high-speed performance.

\section{Conclusion}

We propose RetCompletion, a two-stage method for pluralistic image completion with three key innovations. First, RetNet is applied for the first time in image completion, offering efficient parallel training and recursive inference. Second, we introduce Bi-RetNet, which integrates bidirectional contextual information to enhance image consistency and reconstruction quality. Third, our pixel-wise inference approach significantly reduces inference time, outperforming Transformer-based methods in computational efficiency. Experiments demonstrate that RetCompletion delivers superior image quality over CNN-based methods and achieves comparable results to Transformer approaches, while maintaining faster inference, making it highly suitable for real-time applications.
\bibliographystyle{IEEEbib}
\bibliography{RetCompletion}

\begin{thebibliography}{10}

\bibitem{barnes2009patchmatch}
Connelly Barnes, Eli Shechtman, Adam Finkelstein, and Dan~B Goldman,
\newblock ``Patchmatch: A randomized correspondence algorithm for structural image editing,''
\newblock {\em ACM Trans. Graph.}, vol. 28, no. 3, pp. 24, 2009.

\bibitem{criminisi2004region}
Antonio Criminisi, Patrick P{\'e}rez, and Kentaro Toyama,
\newblock ``Region filling and object removal by exemplar-based image inpainting,''
\newblock {\em IEEE Transactions on image processing}, vol. 13, no. 9, pp. 1200--1212, 2004.

\bibitem{dale2009image}
Kevin Dale, Micah~K Johnson, Kalyan Sunkavalli, Wojciech Matusik, and Hanspeter Pfister,
\newblock ``Image restoration using online photo collections,''
\newblock in {\em 2009 IEEE 12th International Conference on Computer Vision}. IEEE, 2009, pp. 2217--2224.

\bibitem{wan2020bringing}
Ziyu Wan, Bo~Zhang, Dongdong Chen, Pan Zhang, Dong Chen, Jing Liao, and Fang Wen,
\newblock ``Bringing old photos back to life,''
\newblock in {\em proceedings of the IEEE/CVF conference on computer vision and pattern recognition}, 2020, pp. 2747--2757.

\bibitem{CNN3}
Yijun Li, Sifei Liu, Jimei Yang, and Ming-Hsuan Yang,
\newblock ``Generative face completion,''
\newblock in {\em Proceedings of the IEEE conference on computer vision and pattern recognition}, 2017, pp. 3911--3919.

\bibitem{wan2021high}
Ziyu Wan, Jingbo Zhang, Dongdong Chen, and Jing Liao,
\newblock ``High-fidelity pluralistic image completion with transformers,''
\newblock in {\em Proceedings of the IEEE/CVF International Conference on Computer Vision}, 2021, pp. 4692--4701.

\bibitem{zheng2022bridging}
Chuanxia Zheng, Tat-Jen Cham, Jianfei Cai, and Dinh Phung,
\newblock ``Bridging global context interactions for high-fidelity image completion,''
\newblock in {\em Proceedings of the IEEE/CVF Conference on Computer Vision and Pattern Recognition}, 2022, pp. 11512--11522.

\bibitem{li2022mat}
Wenbo Li, Zhe Lin, Kun Zhou, Lu~Qi, Yi~Wang, and Jiaya Jia,
\newblock ``Mat: Mask-aware transformer for large hole image inpainting,''
\newblock in {\em Proceedings of the IEEE/CVF conference on computer vision and pattern recognition}, 2022, pp. 10758--10768.

\bibitem{sun2023retentive}
Yutao Sun, Li~Dong, Shaohan Huang, Shuming Ma, Yuqing Xia, Jilong Xue, Jianyong Wang, and Furu Wei,
\newblock ``Retentive network: A successor to transformer for large language models,''
\newblock {\em arXiv preprint arXiv:2307.08621}, 2023.

\bibitem{CNN4}
Chuanxia Zheng, Tat-Jen Cham, and Jianfei Cai,
\newblock ``Pluralistic image completion,''
\newblock in {\em Proceedings of the IEEE/CVF Conference on Computer Vision and Pattern Recognition}, 2019, pp. 1438--1447.

\bibitem{lugmayr2022repaint}
Andreas Lugmayr, Martin Danelljan, Andres Romero, Fisher Yu, Radu Timofte, and Luc Van~Gool,
\newblock ``Repaint: Inpainting using denoising diffusion probabilistic models,''
\newblock in {\em Proceedings of the IEEE/CVF Conference on Computer Vision and Pattern Recognition}, 2022, pp. 11461--11471.

\bibitem{ho2020denoising}
Jonathan Ho, Ajay Jain, and Pieter Abbeel,
\newblock ``Denoising diffusion probabilistic models,''
\newblock {\em Advances in neural information processing systems}, vol. 33, pp. 6840--6851, 2020.

\bibitem{sun2022length}
Yutao Sun, Li~Dong, Barun Patra, Shuming Ma, Shaohan Huang, Alon Benhaim, Vishrav Chaudhary, Xia Song, and Furu Wei,
\newblock ``A length-extrapolatable transformer,''
\newblock {\em arXiv preprint arXiv:2212.10554}, 2022.

\bibitem{russakovsky2015imagenet}
Olga Russakovsky, Jia Deng, Hao Su, Jonathan Krause, Sanjeev Satheesh, Sean Ma, Zhiheng Huang, Andrej Karpathy, Aditya Khosla, Michael Bernstein, et~al.,
\newblock ``Imagenet large scale visual recognition challenge,''
\newblock {\em International journal of computer vision}, vol. 115, pp. 211--252, 2015.

\bibitem{devlin2018bert}
Jacob Devlin, Ming-Wei Chang, Kenton Lee, and Kristina Toutanova,
\newblock ``Bert: Pre-training of deep bidirectional transformers for language understanding,''
\newblock {\em arXiv preprint arXiv:1810.04805}, 2018.

\bibitem{karras2017progressive}
Tero Karras, Timo Aila, Samuli Laine, and Jaakko Lehtinen,
\newblock ``Progressive growing of gans for improved quality, stability, and variation,''
\newblock {\em arXiv preprint arXiv:1710.10196}, 2017.

\bibitem{liu2018image}
Guilin Liu, Fitsum~A Reda, Kevin~J Shih, Ting-Chun Wang, Andrew Tao, and Bryan Catanzaro,
\newblock ``Image inpainting for irregular holes using partial convolutions,''
\newblock in {\em Proceedings of the European conference on computer vision (ECCV)}, 2018, pp. 85--100.

\bibitem{karras2019style}
Tero Karras, Samuli Laine, and Timo Aila,
\newblock ``A style-based generator architecture for generative adversarial networks,''
\newblock in {\em Proceedings of the IEEE/CVF conference on computer vision and pattern recognition}, 2019, pp. 4401--4410.

\bibitem{suvorov2022resolution}
Roman Suvorov, Elizaveta Logacheva, Anton Mashikhin, Anastasia Remizova, Arsenii Ashukha, Aleksei Silvestrov, Naejin Kong, Harshith Goka, Kiwoong Park, and Victor Lempitsky,
\newblock ``Resolution-robust large mask inpainting with fourier convolutions,''
\newblock in {\em Proceedings of the IEEE/CVF winter conference on applications of computer vision}, 2022, pp. 2149--2159.

\bibitem{liu2020rethinking}
Hongyu Liu, Bin Jiang, Yibing Song, Wei Huang, and Chao Yang,
\newblock ``Rethinking image inpainting via a mutual encoder-decoder with feature equalizations,''
\newblock in {\em Computer Vision--ECCV 2020: 16th European Conference, Glasgow, UK, August 23--28, 2020, Proceedings, Part II 16}. Springer, 2020, pp. 725--741.

\bibitem{CNN5}
Kamyar Nazeri, Eric Ng, Tony Joseph, Faisal~Z Qureshi, and Mehran Ebrahimi,
\newblock ``Edgeconnect: Generative image inpainting with adversarial edge learning,''
\newblock {\em arXiv preprint arXiv:1901.00212}, 2019.

\end{thebibliography}

\end{document}